\documentclass[letterpaper, 10 pt, conference]{ieeeconf}  % Comment this line out if you need a4paper

\IEEEoverridecommandlockouts                              % This command is only needed if 
\let\labelindent\relax
\usepackage{graphics,enumitem} % for pdf, bitmapped graphics files
\usepackage{epsfig} % for postscript graphics files
\usepackage{mathptmx} % assumes new font selection scheme installed
\usepackage{times} % assumes new font selection scheme installed
\usepackage{amsmath} % assumes amsmath package installed
\usepackage{amssymb}  % assumes amsmath package installed
\usepackage{booktabs}
\usepackage{multirow}
\usepackage{algorithm}
\usepackage{algorithmic}
\usepackage{makecell}
\usepackage{caption}
\usepackage{subcaption}
\usepackage{url}
\usepackage{multicol}
\usepackage{pifont}
\usepackage{wrapfig}
\usepackage{cite}
\usepackage{balance}
\usepackage[pagebackref=true,breaklinks=true,colorlinks=true,bookmarks=false,citecolor=blue]{hyperref}
\hypersetup{
colorlinks=true,
linkcolor=blue,
filecolor=magenta,      
urlcolor=blue,
citecolor=blue
}

\title{\LARGE \bf Adaptive Mobile Manipulation for Articulated Objects\\In the Open World}

\author{Haoyu Xiong \\CMU \and Russell Mendonca\\CMU \and Kenneth Shaw\\CMU \and Deepak Pathak\\CMU
\thanks{This work was supported by Carnegie Mellon University.}
}

\begin{document}
\twocolumn[{%
\renewcommand\twocolumn[1][]{#1}%
\maketitle
\begin{center}
    \vspace{-2em}
    \centering
    \captionsetup{type=figure}
    \includegraphics[width=\linewidth]{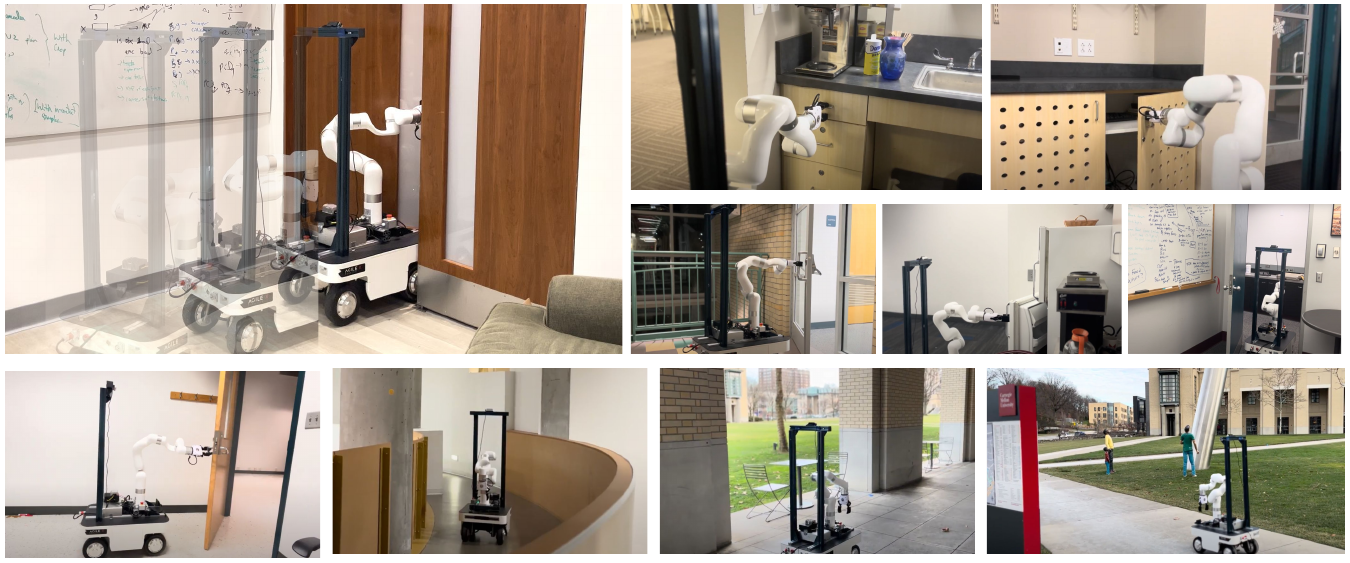}
    \captionof{figure}{\small \textbf{Open-World Mobile Manipulation System}: We use a \textbf{full-stack} approach to operate articulated objects such as real-world doors, cabinets, drawers, and refrigerators in open-ended unstructured environments.  
    }
    \label{fig:teaser}
\end{center}%
}]

\begin{abstract}
Deploying robots in open-ended unstructured environments such as homes has been a long-standing research problem.  
However, robots are often studied only in closed-off lab settings, and prior mobile manipulation work is restricted to pick-move-place, which is arguably just the tip of the iceberg in this area.
In this paper, we introduce Open-World Mobile Manipulation System, a \textbf{full-stack} approach to tackle realistic articulated object operation, e.g. real-world doors, cabinets, drawers, and refrigerators in open-ended unstructured environments.
The robot utilizes an adaptive learning framework to initially learns from a small set of data through behavior cloning, followed by learning from online practice on novel objects that fall outside the training distribution.
We also develop a low-cost mobile manipulation hardware platform capable of safe and autonomous online adaptation in unstructured environments with a cost of around $25,000$ USD.
In our experiments we utilize 20 articulate objects across 4 buildings in the CMU campus. With less than an hour of online learning for each object, the system is able to increase success rate from $50\%$ of BC pre-training to $95\%$ using online adaptation. Video results at \url{https://open-world-mobilemanip.github.io/}.

\end{abstract}

\section{Introduction}

Deploying robotic systems in unstructured environments such as homes has been a long-standing research problem. In recent years, significant progress has been made in deploying learning-based approaches~\cite{whirl, rma, goat, shah2023vint} towards this goal. However, this progress has been largely made independently either in mobility or in manipulation, while a wide range of practical robotic tasks require dealing with both aspects~\cite{mobile-aloha, dobbe, rt1, ruihan}.
The joint study of mobile manipulation paves the way for generalist robots which can perform useful tasks in open-ended unstructured environments, as opposed to being restricted to controlled laboratory settings focused primarily on tabletop manipulation.

However, developing and deploying such robot systems in the \textit{open-world} 
with the capability of handling unseen objects is challenging for a variety of reasons, ranging from the lack of capable mobile manipulator hardware systems to the difficulty of operating in diverse scenarios. Consequently, most of the recent mobile manipulation results end up being limited to pick-move-place tasks\cite{homerobot, trash-sorting, sun2021fully}, which is arguably representative of only a small fraction of problems in this space. Since learning for general-purpose mobile manipulation is challenging, we focus on a restricted class of problems, involving the operation of articulated objects, such as doors, drawers,  refrigerators, or cabinets in open-world environments. This is a common and essential task encountered in everyday life, and is a long-standing problem in the community~\cite{darpa_2, darpa_3, darpa_4, door_old_1, door_old_2, door_old_3, door_old_4}. 
The primary challenge is generalizing effectively across the diverse variety of such objects in unstructured real-world environments rather than manipulating a single object in a constrained lab setup. Furthermore, we also need capable hardware, as opening a door not only requires a powerful and dexterous manipulator, but the base has to be stable enough to balance while the door is being opened and agile enough to walk through. 

 We take a \textbf{full-stack} approach to address the above challenges.
In order to effectively manipulate objects in open-world settings, we adopt a \emph{adaptive learning} approach, where  the robot keeps learning from online samples collected during interaction. Hence even if the robot encounters a new door with a different mode of articulation, or with different physical parameters like weight or friction, it can keep adapting by learning from its interactions. For such a system to be effective, it is critical to be able to learn efficiently, since it is expensive to collect real world samples. The mobile manipulator we use as shown in Figure. \ref{fig:hardware} has a very large number of degrees of freedom, corresponding to the base as well as the arm. A conventional approach for the action space of the robot could be regular end-effector control for the arm and SE2 control for the base to move in the plane. While this is very expressive and can cover many potential behaviors for the robot to perform, we will need to collect a very large amount of data to learn control policies in this space. Given that our focus is on operating  articulated objects, can we structure the action space so that we can get away with needing fewer samples for learning?

Consider the manner in which people typically approach operating articulated objects such as doors. This generally first involves reaching towards a part of the object (such as a handle) and establishing a grasp. We then execute constrained manipulation like rotating, unlatching, or unhooking, where we apply arm or body movement to manipulate the object. In addition to this high-level strategy, there are also lower-level decisions made at each step regarding exact direction of movement, extent of perturbation and amount of force applied. 
Inspired by this, we use a hierarchical action space for our controller, where the high-level action sequence follows the grasp, constrained manipulation strategy. These primitives are parameterized by learned low-level continuous values, which needs to be adapted to operate diverse articulated objects. 
To further bias the exploration of the system towards reasonable actions and avoid unsafe actions during online sampling, we collect a dataset of expert demonstrations on 12 training objects, including doors, drawers and cabinets to train an initial policy via behavior cloning. While this is not very performant on new unseen doors (getting around 50\% accuracy), starting from this policy allows subsequent learning to be faster and safer.

Learning via repeated online interaction also requires capable hardware.  As shown in Figure~\ref{fig:hardware}, we provide a simple and intuitive solution to build a mobile manipulation hardware platform, followed by two main principles: (1) Versatility and agility - this is essential to effectively operate diverse objects with different physical properties in potentially challenging environments, for instance a cluttered office. (2) Affordabiluty and Rapid-prototyping - Assembled with off the shelf components, the system is accessible and can be readily be used by most research labs.

In this paper, we present \textbf{Open-World Mobile Manipulation System}, a \textbf{full stack} approach to tackle the problem of mobile manipulation of realistic articulated objects in the open world.
Efficient learning is enabled by a structured action space with parametric primitives, and by pretraining the policy on a demonstration dataset using imitation learning. Adaptive learning allows the robot to keep learning from self-practice data via online RL. Repeated interaction for autonomous learning requires capable hardware, for which we propose a versatile, agile, low-cost easy to build system.
We introduce a low-cost mobile manipulation hardware platform that offers a high payload, making it capable of repeated interaction with objects, e.g. a heavy, spring-loaded door, and a human-size, capable of maneuvering across various doors and navigating around narrow and cluttered spaces in the open world.
We conducted a field test of 8 novel objects ranging across 4 buildings on a university campus to test the effectiveness of our system, and found adaptive earning boosts success rate from 50\% from the pre-trained policy to 95\% after adaptation.

\section{Related Work}

\paragraph{Adaptive Real-world Robot Learning}
There has been a lot of prior work that studies how robots can acquire new behavior by directly using real-world interaction samples via reinforcement learning using reward~\cite{levine2013guided,levineFDA15, kalashnikov2018qt, kalashnikov2021mt}, and even via unsupervised exploration~\cite{pong2019skewfit, mendonca2023alan}. More recently there have been approaches that use RL to fine-tune policies very efficiently that have been initialized via by imitating demonstrations~\cite{FISH, ROT}. Other methods aim to do so without access to demonstrations on the test objects, and pretrain using other sources of data - either using offline robot datasets~\cite{kumar2022pre}, simulation~\cite{laura-finetune} or human video ~\cite{mendonca2023structured, deft} or a combination of these approaches ~\cite{trash-sorting}.  We operate in a similar setting, without any demonstrations on test objects, and focus on demonstrating RL adaptation on mobile manipulation systems that can be deployed in open-world environments. While prior large-scale industry efforts also investigate this ~\cite{trash-sorting}, we seek to be able to learn much more efficiently with fewer data samples. 

\paragraph{Learning-based Mobile Manipulation Systems. } 

In recent years, the setup for mobile manipulation tasks in both simulated and real-world environments has been a prominent topic of research~\cite{stf-icra-mm, gatech-mm, behavior, habitat, roboturk-mm, mayank-art, opend, wholebody, mobile-aloha, rt-1}.
Notably, several studies have explored the potential of integrating Large Language Models into personalized home robots, signifying a trend towards more interactive and user-friendly robotic systems~\cite{tidybot,saycan,rt-1}. 
While these systems display impressive long horizon capabilities using language for planning, these assume fixed low-level primitives for control. In our work we seek to learn low-level control parameters via interaction. Furthermore, unlike the majority of prior research which predominantly focuses on pick-move-place tasks~\cite{homerobot}, we consider operating articulated objects in unstructured environments, which present an increased level of difficulty.

\paragraph{Door Manipulation} 

The research area of door opening  has a rich history in the robotics community~\cite{door_old_1, door_old_2, door_old_3, door_old_4, old_door_detect}.
A significant milestone in the domain was the DARPA Robotics Challenge (DRC) finals in 2015. The accomplishment of the WPI-CMU team in door opening illustrated not only advances in robotic manipulation and control but also the potential of humanoid robots to carry out intricate tasks in real-world environments~\cite{darpa_2, darpa_3, darpa_4}.
Nevertheless, prior to the deep learning era, the primary impediment was the robots' perception capabilities, which faltered when confronted with tasks necessitating visual comprehension of complex and unstructured environments.
Approaches using deep learning to address vision challenges include Wang et al.~\cite{keypoint_door}, which leverages synthetic data to train keypoint representation for the grasping pose estimation, and 
Qin et, al.~\cite{ucsd_door}, which proposed an end-end point cloud RL framework for sim2real transfer. Another approach is to use simulation to learn policies, using environments such as Doorgym~\cite{doorgym}, which provides a simulation benchmark for door opening tasks. The prospect of large-scale RL combined with sim-to-real transfer holds great promise for generalizing to a diverse range of doors in real-world settings~\cite{doorgym, uiuc_door, ucsd_door}. However, one major drawback is that the system can only generalize to the space of assets already present while training in the simulation. Such policies might struggle when faced with a new unseen door with physical properties, texture or shape different from the training distribution. Our approach can keep on learning via real-world samples, and hence can learn to adapt to difficulties faced when operating new unseen doors.

\section{Adaptive Learning Framework}

In this section, we describe our algorithmic framework for training robots for adaptive mobile manipulation of everyday articulated objects. 
To achieve efficient learning, we use a structured hierarchical action space. This uses a fixed high-level action strategy and learnable low-level control parameters. Using this action space, we initialize our policy via behavior cloning (BC) with a diverse dataset of teleoperated demonstrations. This provides a strong prior for exploration and decreases the likelihood of executing unsafe actions. 
However, the initialized BC policy might not generalize to every unseen object that the robot might encounter due to the large scope of variation of objects in open-world environments.
To address this, we enable the robot to learn from the online samples it collects to continually learn and adapt. 
We describe the continual learning process as well as design considerations for online learning.

\begin{figure}[t!]
  \centering
  \vspace{-.25in}
  \includegraphics[width=\linewidth]{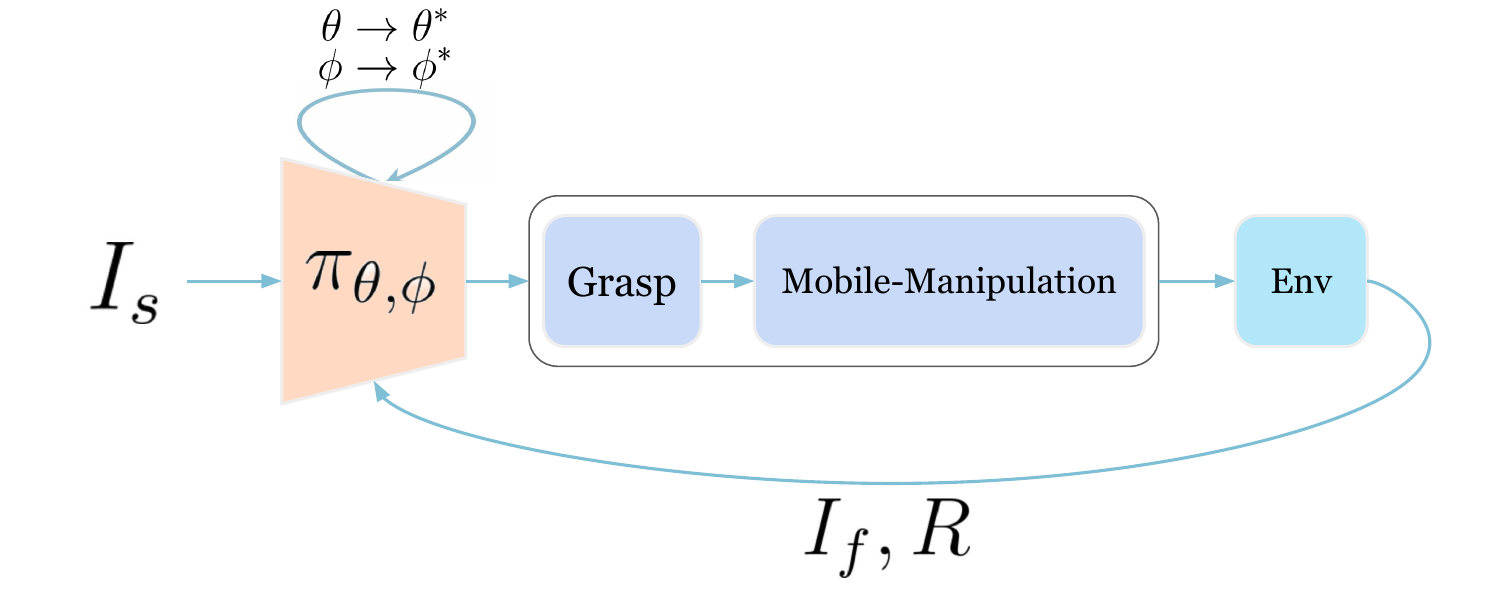}
  \caption{\textbf{Adaptive Learning Framework}: The policy outputs low-level parameters for the grasping primitive, and chooses a sequence of manipulation primitives and their parameters.}
  \label{fig:hardware}
\end{figure}

\subsection{Action Space}

For greater learning efficiency, we use a parameterized primitive action space. Concretely, we assume access to a grasping primitive $G(.)$ parameterized by $g$. We also have a constrained mobile-manipulation primitives $M(.)$, where primitive  $M(.)$ takes two parameters, a discrete parameter $C$ and a continuous parameter $c$.
Trajectories are executed in an open-loop manner, a grasping primitive followed by a sequence of N constrained mobile-manipulation primitives:

\begin{equation*}
    \{I_s, G(g),\{ M(C_i, c_i) \}_{i=1}^{N}, I_f, R\}
\end{equation*}

where $I_s$ is the initial observed image, $G(g)$,  $M(C_i,c_i))$ denote the parameterized grasp and constrained manipulation primitives respectively, $I_f$ is the final observed image, and $r$ is the reward for the trajectory. While this structured space is less expressive than the full action space, it is large enough to learn effective strategies for the everyday articulated objects we encountered, covering 20 different doors, drawers, and fridges in open-world environments. The key benefit of the structure is that it allows us to learn from very few samples, using only on the order of 20-30 trajectories. 
We describe the implementation details of the primitives in section \ref{subsec:prim_impl}.

\begin{algorithm}[t!]
\caption{Adaptive Learning}
\label{alg:method}
\begin{algorithmic}[1]

\REQUIRE Grasping primitive $G(.)$ taking parameter $g$
\REQUIRE Constrained manipulation primitives $ M(.)$, taking parameter C and c.

\STATE Initialize primitive classifier $\pi_{\phi}(\{C_i \}_{i=1}^{N}|I)$
\STATE Initialize conditional action policy\\ $\pi_{\theta}(g, \{c_i \}_{i=1}^{N}| I, \{C_i \}_{i=1}^{N})$

\STATE Collect a dataset $D$ of expert demos \\$\{ I, g, \{C_i \}_{i=1}^{N}, \{c_i \}_{i=1}^{N}\} $

\STATE Train $\pi_\phi$ and $\pi_\theta$ on $D$ using Imitation Learning \ref{eq:BC}
\FOR {online RL iteration 1:N}
    \STATE Given image $I_s$, sample $\{C_i \}_{i=1}^{N} \sim \pi_{\phi}(.|I_s)$, \\ sample $(g, \{c_i \}_{i=1}^{N}) \sim \pi_{\theta}(.|I_s)$
    \STATE Execute trajectory $\{ G(g),\{ M(C_i, c_i) \}_{i=1}^{N}\}$, \\observe reward $R$
    \STATE Update policies $\pi_\phi$ and $\pi_\theta$ using RL (Eqs. \ref{eq:overall}, \ref{eq:RL}, \ref{eq:BC})
\ENDFOR

\end{algorithmic}
\vspace{-0.05in}
\label{algo:action_space}
\end{algorithm}

\subsection{Adaptive Learning}

Given an initial observation image $I_s$, we use a classifier $\pi_\phi(\{C_i \}_{i=1}^{N}|I)$ to predict the a sequence of N discrete parameters $\{C_i \}_{i=1}^{N}$ for constrained mobile-manipulation, and a conditional policy network $\pi_\theta(g, \{c_i \}_{i=1}^{N} | I, \{C_i \}_{i=1}^{N})$  which produces the continuous parameters of the grasping primitive and a sequence of N constrained mobile-manipulation primitives. The robot executes the parameterized primitives one by one in an open-loop manner. 

\subsubsection{Imitation}
We start by initializing our policy using a small set of expert demonstrations via behavior cloning. The details of this dataset are described in section \ref{subsec:pretraining_dataset}. The imitation learning objective is to learn policy parameters $\pi_{\theta, \phi}$ that maximize the likelihood of the expert actions. Specifically, given a dataset of image observations $I_s$, and corresponding actions $\{g, \{C_i \}_{i=1}^{N}, \{c_i \}_{i=1}^{N}\}$, the imitation learning objective is: 

\begin{equation}
\max_{\phi, \theta} \left[ \log \pi_{\phi}(\{C_i \}_{i=1}^{N} \,|\, I_s) + \log \pi_{\theta}(g, \{c_i \}_{i=1}^{N} \,|\, \{C_i \}_{i=1}^{N}, I_s) \right]
\end{equation}

\subsubsection{Online RL}

The central challenge we face is operating new articulated objects that fall outside the behavior cloning training data distribution. To address this, we enable the policy to keep improving using the online samples collected by the robot. This corresponds to maximizing the expected sum of rewards under the policy : 
\begin{equation}
\label{eq:BC}
    \max_{\theta, \phi} \mathbb{E}_{\pi_{\theta, \phi}}\left[\sum_{t=0}^{T} r(s_t, a_t)\right]
\end{equation}

Since we utilize a highly structured action space as described previously, we can optimize this objective using a fairly simple RL algorithm. Specifically we use the REINFORCE objective \cite{reinforce}: 

\begin{align}
    &\nabla_{\theta, \phi} J(\theta, \phi) = \mathbb{E}_{\pi_{\theta, \phi}}\left[\sum_{t=0}^{T} \nabla_{\theta} \log \pi(a_t|s_t) \cdot r_t\right] \\ 
    \label{eq:RL}
    &= \mathbb{E}_{\pi_{\phi, \theta}}\left[ (\nabla_{\phi} \log \pi_{\phi}(C_i|I) + \nabla_{\theta} \log \pi_{\theta}(g, c_i|C_i, I))  \cdot R\right] 
\end{align}

where $R$ is the reward provided at the end of trajectory execution. Note that we only have a single time-step transition, all actions are determined from the observed image $I_s$, and executed in an open-loop manner. Further details for online adaptation such as rewards, resets and safety are detailed in section \ref{subsec:auto_ada}.

\subsubsection{Overall Finetuning Objective}
To ensure that the policy doesn't deviate too far from the initialization of the imitation dataset, we use a weighted objective while finetuning, where the overall loss is  : 

\vspace{-0.1in}
\begin{equation}
\label{eq:overall}
   \mathcal{L}_\text{overall} = \mathcal{L}_\text{online} + \alpha*\mathcal{L}_\text{offline}
\end{equation}

where loss on online sampled data is optimized via Eq.\ref{eq:RL} and loss on the batch of offline data is optimized via BC as in Eq.\ref{eq:BC}. We use equal sized batches for online and offline data while performing the update. 

\begin{figure}[t!]
  \centering
  \vspace{-.25in}
  \includegraphics[width=0.75\linewidth]{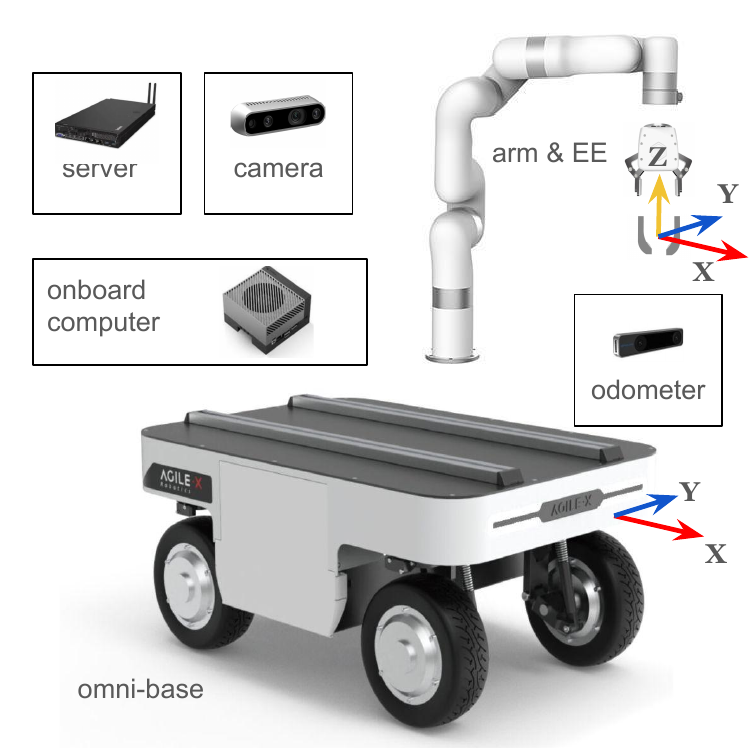}
  \caption{\textbf{Mobile Manipulation Hardware Platform:} Different components in the mobile manipulator hardware system. Our design is low-cost and easy-to-build with off-the-shelf components}

  \label{fig:hardware}
\end{figure}

\begin{figure*}[!ht]
    \centering
    \includegraphics[width=\textwidth]{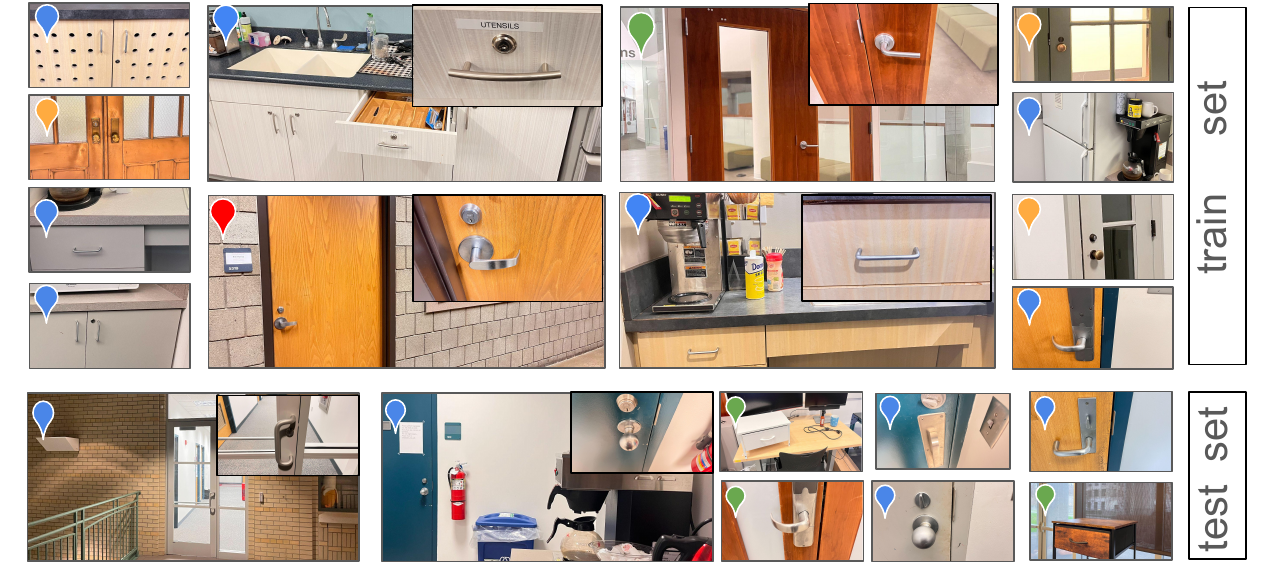}
    \caption{\small \textbf{Articulated Objects}: Visualization of the 12 training and 8 testing objects used, with location indicators corresponding to the buildings in the map below. The training and testing objects are significantly different from each other, in terms of different visual appearances, different modes of articulation, or different physical parameters, e.g. weight or friction.}
    \label{fig:all_doors}
\end{figure*}

\begin{figure}[t]
  \begin{center}
    \includegraphics[width=1\linewidth]{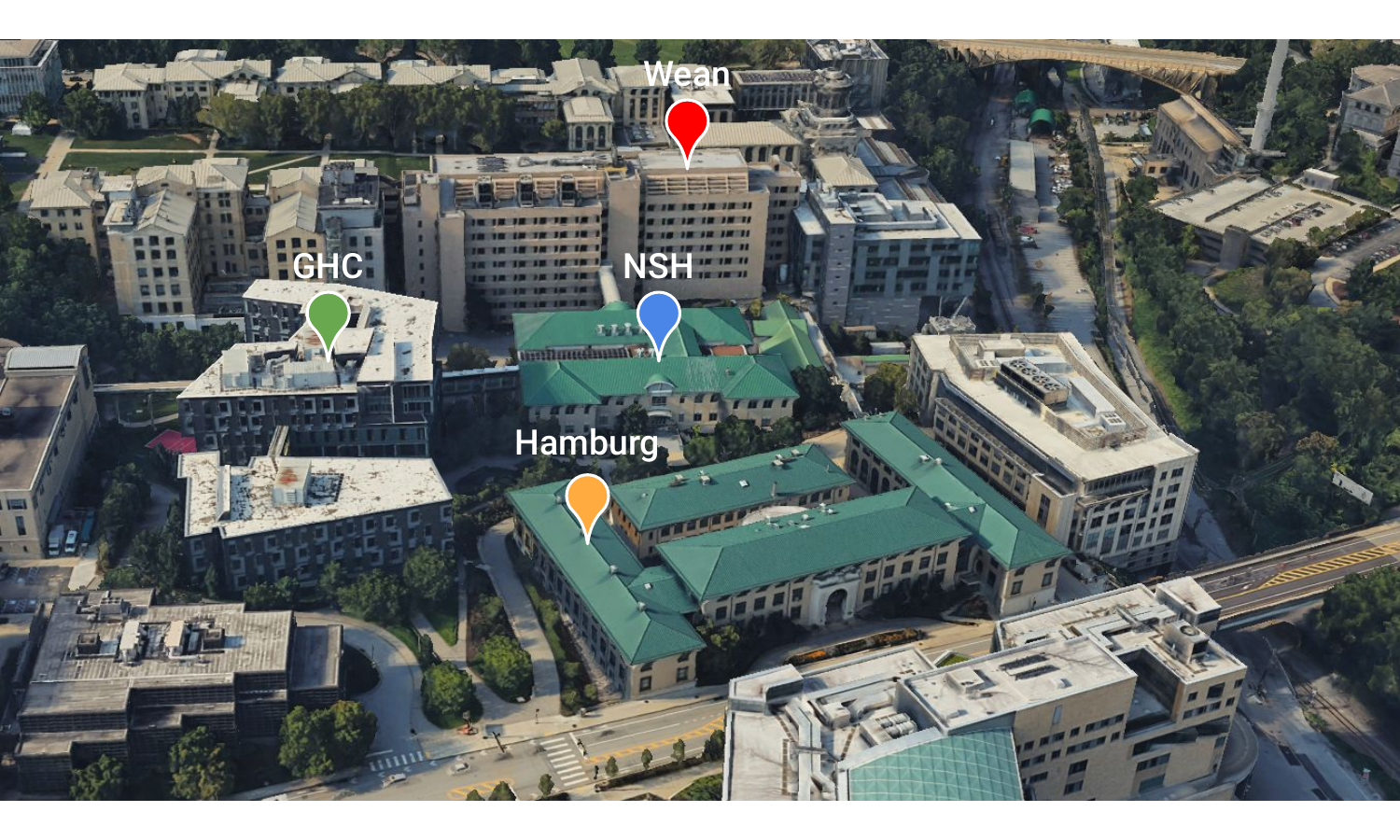} 
  \end{center}
  \caption{\textbf{Field Test on CMU Campus}: The system was evaluated on articulated objects from across four distinct buildings on the Carnegie Mellon University campus. }
  \label{fig:map}
\end{figure}

\section{Open-world Mobile Manipulation Systems}

In this section, we describe details of our \textit{full-stack} approach encompassing hardware, action space for efficient learning, the demonstration dataset for initialization of the policy and crucially details of autonomous, safe execution with rewards.  
This enables our mobile manipulation system to adaptively learn in open-world environments, to manipulate everyday articulated objects like cabinets, drawers, refrigerators, and doors.

\subsection{Hardware} 

The transition from tabletop manipulation to mobile manipulation is challenging not only from algorithmic studies but also from the perspective of hardware. 
In this project, we provide a simple and intuitive solution to build a mobile manipulation the hardware platform. Specifically, our design addresses the following challenges - 
\begin{itemize}
    \item \emph{Versatility and agility}:  Everyday articulated objects like doors have a wide degree of variation of physical properties, including weight, friction and resistance. To successfully operate these, the platform must offer high payload capabilities via a strong arm and base. Additionally, we sought to develop a human-sized, agile platform capable of maneuvering across various real-world doors and navigating unstructured and narrow environments, such as cluttered office spaces.
    \item \emph{Affordability and Rapid-Prototyping}: The platform is designed to be low-cost for most robotics labs and employs off-the-shelf components. This allows researchers to quickly assemble the system with ease, allowing the possibility of large-scale open-world data collection in the future. 

\end{itemize}

We show the different components of the hardware system in Figure~\ref{fig:hardware}. Among the commercially available options, we found the Ranger Mini 2 from AgileX to be an ideal choice for robot base due to its stability, omni-directional velocity control, and high payload capacity. The system uses an xArm for manipulation, which is an effective low-cost arm with a high payload (5kg), and is widely accessible for research labs.
The system uses a Jetson computer to support real-time communication between sensors, the base, the arm, as well as a server that hosts large models. 
We use a D435 Intel Realsense camera mounted on the frame to collect RGBD images as ego-centric observations and a T265 Intel Realsense camera to provide visual odometry which is critical for resetting the robot when performing trials for RL.  
The gripper is equipped with a 3d-printed hooker and an anti-slip tape to ensure a secure and stable grip. 
The overall cost of the entire system is around $25,000$ USD, making it an affordable solution for most robotics labs.

We compare key aspects of our modular platform with that of other mobile manipulation platforms in Table ~\ref{tab:specsheet}.This comparison highlights advantages of our system such as cost-effectiveness, reactivity, ability to support a high-payload arm, and a base with omnidirectional drive.

\subsection{Primitive Implementation}
\label{subsec:prim_impl}
In this subsection, we describe the implementation details of our parameterized primitive action space.

\subsubsection{Grasping}
Given the RGBD image of the scene obtained from the realsense camera, we use off-the-shelf visual models \cite{detic, sam} to obtain the mask of the door and handle given just text prompts. Furthermore, since the door is a flat plane, we can estimate the surface normals of the door using the corresponding mask and the depth image. This is used to move the base close to the door and align it to be perpendicular, and also to set the orientation angle for grasping the handle. The center of the 2d mask of the handle is projected into 3d coordinates using camera calibration, and this is the nominal grasp position. The low-level control parameters to the grasping primitive indicate an offset for this position at which to grasp. 
This is beneficial since depending on the type of handle the robot might need to reach a slightly different position which can be learned via the low-level continuous valued parameters.

\begin{table*}[t!]
    \centering
    \resizebox{\linewidth}{!}{
    \begin{tabular}{lcccccc}
        \toprule

        \multicolumn{5}{c}{Hardware features comparison} \\
        \midrule
        & Arm payload & DoF arm & omni-base & footprint & base max speed & price \\
        Stretch RE1~\cite{ruihan} & 1.5kg & 2 & \ding{53} & 34 cm, 33 cm & 0.6 m/s & 20k USD\\ 
        Go$1$-air + WidowX 250s~\cite{wholebody} & 0.25kg & 6 & \ding{51} & 59 cm, 22 cm & 2.5 m/s & 10k USD \\
        Franka + Clearpath Ridgeback ~\cite{frankaclearpath}& 3kg & 7 & \ding{51}  & 96 cm, 80 cm  & 1.1 m/s & 75k USD \\
        Franka + Omron LD-60 ~\cite{sayplan} & 3kg & 7 & \ding{53} & 70 cm, 50 cm & 1.8 m/s & 50k USD \\
        Xarm-6 + Agilex Ranger mini 2 (ours) & 5kg & 6 & \ding{51}  & 74 cm, 50 cm & 2.6 m/s & 25k USD\\

        \midrule

    \end{tabular}
    }
    \caption{Comparison of different aspects of popular hardware systems for mobile manipulation}
    \label{tab:specsheet}
\end{table*}

\subsubsection{Constrained Mobile-Manipulation}

We use velocity control for the robot arm end-effector and the robot base. With a 6dof arm and 3dof motion for the base (in the SE2 plane), we have a 9-dimensional vector - 
\begin{align*}
    \text{Control} &: \quad (v_x, v_y, v_z, v_\text{yaw}, v_\text{pitch}, v_\text{roll}, V_\text{x}, V_\text{y}, V_\omega ) 
\end{align*}
Where the first 6 dimensions correspond to control for the arm, and the last three are for the base. The primitives we use impose contraints on this space as follows - 
\begin{align*}
    \text{Unlock} &: \quad (0,0,v_z,v_\text{yaw},0,0,0,0,0) \\
    \text{Rotate} &: \quad (0,0,0,v_\text{yaw},0,0,0,0,0) \\
    \text{Open}   &: \quad (0,0,0,0,0,0,V_\text{x},0,0) \\
\end{align*}

For control, the policy outputs an index corresponding to which primitive is to executed, as well as the corresponding low-level parameters for the motion. The low-level control command is continuous valued from -1 to 1 and executed for a fixed duration of time.  The sign of the parameters dictates the direction of the velocity control, either clockwise or counter-clockwise for unlock and rotate, and forward or backward for open.

\subsection{Pretraining Dataset}
\label{subsec:pretraining_dataset}

The articulated objects we consider in this project consist of three rigid parts: a base part, a frame part, and a handle part. This covers objects such as doors, cabinets, drawers and fridges. The base and frame are connected by either a revolute joint (as in a cabinet) or a prismatic joint (as in a drawer). The frame is connected to the handle by either a revolute joint or a fixed joint.
We identify four major types of the articulated objects, which relate to the type of handle, and the joint mechanisms. 
Handle articulations commonly include levers (Type A) and knobs (Type B). For cases where handles are not articulated, the body-frame can revolve about a hinge using a revolute joint (Type C), or slide back and forth along a prismatic joint, for example, drawers (Type D). While not exhaustive, this categorization covers a wide variety of everyday articulated objects a robot system might encounter. 
To provide generalization benefits in operating unseen novel articulated objects, we first collect a offline demonstration dataset. We include 3 objects from each category in the BC training dataset, collecting 10 demonstrations for each object, producing a total of 120 trajectories.

We also have 2 held-out testing objects from each category for generalization experiments. The training and testing objects differ significantly in visual appearance (eg. texture, color), physical dynamics (eg. if spring-loaded), and actuation (e.g. the handle joint might be clockwise or counter-clockwise). We include visualizations of all objects used in train and test sets in Fig. \ref{fig:all_doors}, along with which part of campus they are from as visualized in Fig. \ref{fig:map}.

\subsection{Autonomous and Safe Online Adaptation}
\label{subsec:auto_ada}

The key challenge we face is operating with new objects that fall outside the BC training domain. To address this, we develop a system capable of fully autonomous Reinforcement Learning (RL) online adaptation. 
In this subsection, we demonstrate the details of the autonomy and safety of our system.

\subsubsection{Safety Aware Exploration} 
It is crucial to ensure that the actions the robot takes for exploring are safe for its hardware, especially since it is interacting with objects under articulation constraints. Ideally, this could be addressed for dynamic tasks like door opening using force control. However, low-cost arms like the xarm-6 we use do not support precise force sensing. 
For deploying our system, we use a safety mechanism based which reads the joint current during online sampling. If the robot samples an action that causes the joint current to meet its threshold, we terminate the episode and reset the robot, to prevent the arm from potentially damaging itself, and also provide negative reward to disincentivize such actions.

\subsubsection{Reward Specification} 
In our main experiments, a human operator provides rewards- with +1 if the robot succesfully opens the doors, 0 if it fails, and -1 if there is a safety violation. This is feasible since the system requires very few samples for learning. For autonomous learning however, we would like to remove the bottleneck of relying on humans to be present in the loop. We investigate using large vision language models as a source of reward. Specifically, we use CLIP~\cite{clip} to compute the similarity score between two text prompts and the image observed after robot execution. The two prompts we use are -  
\textit{"door that is closed"} and \textit{"door that is open"}. 
We compute the similarity score of the final observed image and each of these prompts and assign a reward of +1 if the image is closer to the prompt indicating the door is open, and 0 in the other case. If a safety protection is triggered the reward is -1.

\subsubsection{Reset Mechanism} 

The robot employs visual odometry, utilizing the T265 tracking camera mounted on its base, enabling it to navigate back to its initial position. 
At the end of every episode, the robot releases its gripper, and moves back to the original SE2 base position, and takes an image of $I_f$ for computing reward. We then apply a random perturbation to the SE2 position of the base so that the policy learns to be more robust.
Furthermore, if the reward is 1, where the door is opened, the robot has a scripted routine to close the door.

\begin{figure*}[!ht]
  \begin{center}
    {\includegraphics[width=1\linewidth]{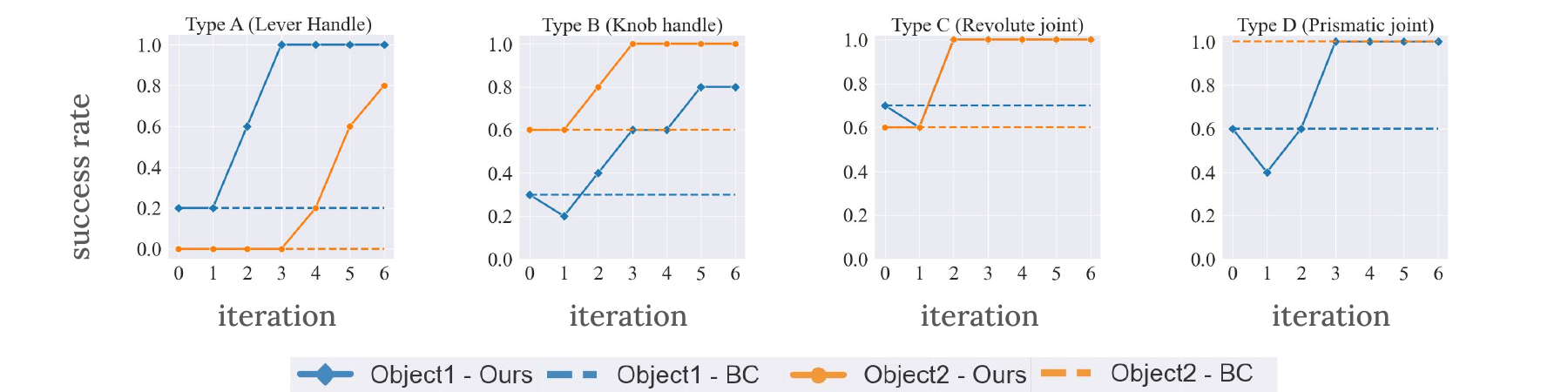}} 
    \vspace{1.0mm}
    
    \vspace{1.0mm}
    
    \captionof{figure}{
    \textbf{Online Improvement}: Comparison of our approach to the imitation policy on 4 different categories of articulated objects, each consisting of two different objects. Our adaptive approach is able to improve in performance, while the imitation policy has limited generalization. 
    }
    \label{fig:adapt}
  \end{center}
\end{figure*}

\section{Results}
We conduct an extensive field study involving $12$ training objects and $8$ testing objects across four distinct buildings on the Carnegie Mellon University campus to test the efficacy of our system. In our experiments, we seek to answer the following questions:

\begin{enumerate}
    \item Can the system improve performance on unseen objects via online adaptation across diverse object categories?
    \item How does this compare to simply using imitation learning on provided demonstrations?
    \item Can we automate providing rewards using off-the-shelf vision-language models?
    \item How does the hardware design compare with other platforms?
\end{enumerate}

\begin{table}
    \centering
    \resizebox{\linewidth}{!}{
    \begin{tabular}{lccc}
        \toprule
        \multicolumn{4}{c}{CLIP-reward comparison} \\
        \midrule
              & BC-0  & Adapt-GT & Adapt-CLIP \\
        Success Rate A1 (lever) & 20\% & 100\% & 80\% \\ 
        Success Rate B1 (knob)  & 30\% & 80\% & 80\% \\ 
        \midrule
    \end{tabular}
}
    \caption{ In this table, we present improvements in online adaptation with CLIP reward.}
    \label{tab:clip}
\end{table}

\subsection{Online Improvement }

\subsubsection{Diverse Object Category Evaluation}:
We evaluate our approach on 4 categories of held-out articulated objects. As described in section \ref{subsec:pretraining_dataset}, these are determined by handle articulation and joint mechanisms. This categorization is based on types of handles, including levers (type A) and knobs (type B), as well as joint mechanisms including revolute (type C) and prismatic (type D) joints. We have two test objects from each category. We report continual adaptation performance in Fig. \ref{fig:adapt} over 5 iterations of fine-tuning using online interactions, starting from the behavior cloned initial policy. Each iteration of improvement consists of 5 policy rollouts, after which the model is updated using the loss in Equation \ref{eq:overall}.

From Fig. \ref{fig:adapt}, we see that our approach improves the average success rate across all objects from 50 to 95 percent. Hence, continually learning via online interaction samples is able to overcome the limited generalization ability of the initial behavior cloned policy.  The adaptive learning procedure is able to learn from trajectories that get high reward, and then change its behavior to get higher reward more often. In cases where the BC policy is reasonably performant, such as Type C and D objects with an average success rate of around 70 percent, RL is able to perfect the policy to 100 percent performance. Furthermore, RL is also able to learn how to operate objects even when the initial policy is mostly unable to perform the task. This can be seen from the Type A experiments, where the imitation learning policy has a very low success rate of only 10 percent, and completely fails to open one of the two doors. With continual practice, RL is able to achieve an average success of 90 percent. This shows that RL can explore to take actions that are potentially out of distribution from the imitation dataset, and learn from them, allowing the robot to learn how to operate novel unseen articulated objects.

\begin{table}
    \centering
    \resizebox{\linewidth}{!}{
    \begin{tabular}{lcccc}
        \toprule
        \multicolumn{5}{c}{Action-Replay Comparison} \\
        \midrule
                & KNN-open  & KNN-close & BC-0 & Adapt-GT \\
        Success Rate B1 (knob) & 10\% & 0\% & 30\% & 80\% \\ 
        Success Rate A2 (lever) & 0\% & 0\% & 0\% & 80\% \\ 
        \midrule
    \end{tabular}
    }
    \caption{We compare the performance of our adaptation policies and initialized BC policies with KNN baselines.}
    \label{tab:knn}
\end{table}

\subsubsection{Action-replay baseline}:
There is also another very simple approach for utilizing a dataset of demonstrations for performing a task on a new object. This involves replaying trajectories from the closest object in the training set. This closest object can be found using k-nearest neighbors with some distance metric. This approach is likely to perform well especially if the distribution gap between training and test objects is small, allowing the same actions to be effective. We run this baseline for two objects that are particularly hard for behavior cloning, one each from Type A and B categories  (lever and knob handles respectively). The distance metric we use to find the nearest neighbor in the training set is euclidean distance of the the CLIP encoding of observed images.
We evaluate this baseline both in an open-loop and closed-loop manner. In the former case, only the first observed image is used for comparison and the entire retrieved action sequence is executed, and in the latter we search for the closest neighbor after every step of execution and perform the corresponding action. 
From Table ~\ref{tab:knn} we see that this approach is quite ineffective, further underscoring the distribution gap between the training and test objects in our experiments. 

\subsubsection{Autonomous reward via VLMs}

We investigate whether we can replace the human operator with an automated procedure to provide rewards. The reward is given by computing the similarity score between the observed image at the end of execution, and two text prompts, one of which indicate that the door is open, and the other that says the doors is closed, as described in section \ref{subsec:auto_ada}. 

\begin{wrapfigure}{r}{0.25\textwidth}
    \vspace{-0.20in}
    \centering
    \includegraphics[width=\linewidth]{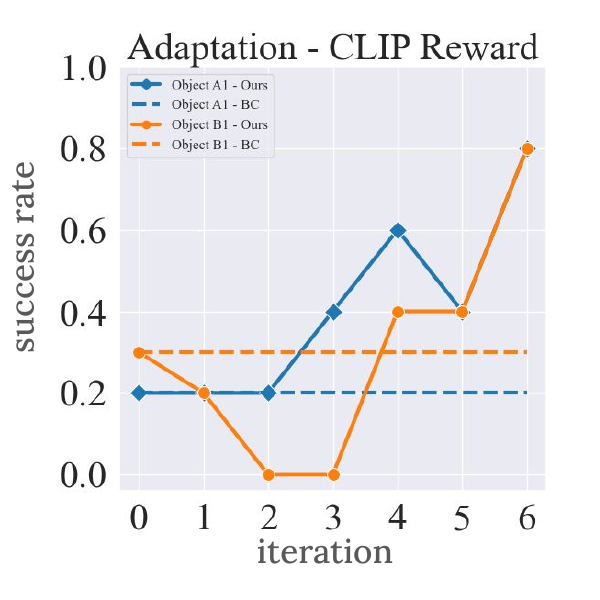}
    \caption{\small \textbf{Online Adaptation with CLIP reward.} Adaptive learning using rewards from CLIP, instead of a human operator, showing our system can operate autonomously.}
    \label{fig:clip}
\end{wrapfigure}

As with the action-replay baseline, we evalute this on two test doors, on each from the handle and knob categories. From Table~\ref{tab:clip}, we see that online adaptation with VLM reward achieves a similar performance as using ground-truth human-labeled reward, with an average of 80 percent compared to 90 percent. We also report the performance after every iteration of training in Fig.~\ref{fig:clip}. Removing the need for a human operator to be present in the learning loop opens up the possiblity for autonomous training and improvement. 

\subsection{Hardware Teleop Strength}

\begin{table}[h!]
    \centering
    \resizebox{0.6\linewidth}{!}{
    \begin{tabular}{lcc}
        \toprule

        \multicolumn{3}{c}{Expert teleoperation success rate} \\
        \midrule
              &lever B& knob A   \\
        Stretch RE1 & $0/5$ & $0/5$ \\
        Ours & $5/5$ & $5/5$ \\

        \midrule

    \end{tabular}
    }
    \caption{\small Human expert teleoperation success rate using stretch and our system for opening doors}
    \label{tab:teleop}
\end{table}

In order to successfully operate various doors the robot needs to be strong enough to open and move through them. We empirically compare against a different popular mobile manipulation system, namely the Stretch RE1 (Hello Robot). We test the ability of the robots to be teleoperated by a human expert to open two doors from different categories, specifically lever and knob doors. Each object was subjected to five trials. As shown is Table~\ref{tab:teleop}, the outcomes of these trials revealed a significant limitation of the Stretch RE1: its payload capacity is inadequate for opening a real door, even when operated by an expert, while our system succeeds in all trials.  

\section{Conclusion}

We present a full-stack system for adaptive learning in open world environments to operate various articulated objects, such as doors, fridges, cabinets and drawers. The system is able to learn from very few online samples since it uses a highly structured action space, which consists of a parametric grasp primitive, followed by a sequence of parametric constrained mobile manipulation primitives. The exploration space is further structured via a demonstration dataset on some training objects. Our approach is able to improve performance from about 50 to 95 percent across 8 unseen objects from 4 different object categories, selected from buildings across the CMU campus. The system can also learn using rewards from VLMs without human intervention, allowing for autonomous learning. We hope to deploy such mobile manipulators to continuously learn a broader variety of tasks via repeated practice. 

\balance
\section*{ACKNOWLEDGMENT}
We thank Shikhar Bahl, Tianyi Zhang, Xuxin Cheng, Shagun Uppal, and Shivam Duggal for the helpful discussions. This work was supported in part by CMU-AIST Bridge project, AFOSR research grant FA9550-23-1-0747 and Sony faculty award.

{
\bibliographystyle{IEEEtran}
\bibliography{paperpile, bib}

}

\end{document}